\documentclass{article} 
\usepackage{iclr2026_conference,times}

\usepackage{amsmath,amsfonts,bm}

\def\eqref#1{equation~\ref{#1}}

\def\1{\bm{1}}

\DeclareMathAlphabet{\mathsfit}{\encodingdefault}{\sfdefault}{m}{sl}
\SetMathAlphabet{\mathsfit}{bold}{\encodingdefault}{\sfdefault}{bx}{n}

\usepackage{hyperref}
\usepackage{url}
\usepackage{fancyvrb}
\usepackage{xcolor}
\usepackage{booktabs}
\usepackage{multirow}
\usepackage[dvipsnames]{xcolor}
\usepackage{amssymb}
\usepackage[most]{tcolorbox} 
\usepackage{enumitem}
\usepackage{graphicx}
\usepackage{fontawesome5}

\tcbset{
  verticalbox/.style={
    enhanced,
    colback=#1!6,
    colframe=#1!60!black,
    boxrule=0.6pt,
    arc=2mm,
    left=1.5mm,right=1.5mm,top=1mm,bottom=1mm,
    before skip=0.6\baselineskip,
    after skip=0.6\baselineskip,
  }
}

\title{ViTaB-A: Evaluating Multimodal Large Language Models on Visual Table Attribution}

\author{
  \begin{tabular}{c}
    \\
    {\large\bfseries Yahia Alqurnawi\thanks{equal contribution.} \quad
    Preetom Biswas\footnotemark[1] \quad
    Anmol Rao\footnotemark[1]} \\[8pt]
    {\large\bfseries Tejas Anvekar \quad Chitta Baral \quad Vivek Gupta} \\[10pt]
    {\normalsize\normalfont School of Computing and Augmented Intelligence} \\
    {\normalsize\normalfont Arizona State University} \\
    {\normalsize\normalfont Tempe, AZ 85281, USA} \\
    {\normalsize\normalfont\texttt{\{yalqurna,pbiswa11,arao75,tanvekar,cbaral,vgupt140\}@asu.edu}}\\
    \\
    {\bfseries     
    \faGlobe~\href{https://preetom1905011.github.io/ViTaB-A/}{Project page} \quad\faDatabase~\href{https://huggingface.co/datasets/raoanmol/ViTaB-A}{Dataset} \quad
    \faGithub~\href{https://github.com/Yahialqur/ViTAB-A}{Code}
    }
  \end{tabular}
}

\iclrfinalcopy 
\begin{document}

\maketitle

\begin{abstract}
Multimodal Large Language Models (mLLMs) are often used to answer questions in structured data such as tables in Markdown, JSON, and images. While these models can often give correct answers, users also need to know where those answers come from. In this work, we study structured data attribution/citation, which is the ability of the models to point to the specific rows and columns that support an answer. We evaluate several mLLMs across different table formats and prompting strategies. Our results show a clear gap between question answering and evidence attribution. Although question answering accuracy remains moderate, attribution accuracy is much lower, near random for JSON inputs, across all models. We also find that models are more reliable at citing rows than columns, and struggle more with textual formats than images. Finally, we observe notable differences across model families. Overall, our findings show that current mLLMs are unreliable at providing fine-grained, trustworthy attribution for structured data, which limits their usage in applications requiring transparency and traceability. 
\end{abstract}

\section{Introduction}

Multimodal Large Language Models (mLLMs) are increasingly used to answer questions over structured data. In practice, users rely on these models to read tables, extract values, compare entries, and summarize records across formats such as Markdown tables, JSON files, and document images. Prior work shows that mLLMs can often answer questions about structured inputs with reasonable accuracy, making them attractive as general-purpose data assistants \citep{fang2024large, liu2024llms}.

However, answering a question correctly is often not enough. In many real-world scenarios, users also want to know where an answer came from. For example, if a model reports that a company’s revenue increased in a given year, a natural follow-up is which row and which column in the table support this claim. In current systems, this step is frequently unreliable: models may produce a correct answer while failing to identify the specific part of the table that justifies it. We demonstrate this gap empirically, showing that question answering accuracy remains relatively high while attribution or citation accuracy is substantially lower across models and prompting strategies (Section \ref{sec:attribution_results}).

In this paper, we study structured data attribution-the ability of mLLMs not only to generate correct answers, but to localize the rows and columns in the input data that support those answers. We evaluate attribution across three common table representations-Markdown, JSON, and images-using multiple model families and prompting strategies.

Our study is motivated by the observation that answer accuracy and attribution accuracy are distinct capabilities. Prior work shows that models can often arrive at correct answers without being fully grounded in the underlying evidence, particularly when partial cues or broadly relevant context are sufficient \citep{bohnet2022attributed, radevski2025synthesizing}. Our results provide concrete evidence of this disparity in structured data settings: across all evaluated models, question answering accuracy remains around 48–55\%, while attribution accuracy is dramatically lower-often below 30\%, and near random for JSON inputs (Section~\ref{sec:attribution_results}).

These findings align with broader evidence that attribution and citation remain challenging for language models. Prior work on hallucination shows that models often generate confident but ungrounded outputs, including incorrect or fabricated references \citep{huang2025survey}. Even when explicitly prompted to cite sources, models frequently produce vague or incorrect attributions \citep{gao2023enabling}. Fine-grained structured attribution benchmarks-such as TabCite-have been introduced to assess models’ ability to locate relevant table structures (e.g., rows and columns), highlighting that precise localization remains an open challenge~\citep{mathur2024matsa}.

Across our experiments, we observe several consistent patterns. First, except in JSON settings, models are substantially better at identifying the correct row than the correct column, suggesting persistent difficulty with fine-grained field-level localization. Third, attribution is more reliable when tables are presented as images than when they are provided in textual formats such as Markdown or JSON. Finally, we observe notable differences across model families, indicating that architectural choices influence attribution behavior.

These limitations are particularly concerning in domains such as finance, healthcare, and law, where systems must support auditability and traceability. In such settings, it is not sufficient to provide a plausible answer; outputs must be traceable to specific data fields. Our results show that current mLLMs are unreliable at providing this level of fine-grained traceability, even when their answers appear correct (Section~\ref{sec:attribution_results}).

Finally, we summarize our contributions as follows:
\begin{itemize}
    \item We propose ViTaB-A, an exhaustive benchmark for assessing mLLMs on Visual Table Attribution tasks across modalities (text, JSON, rendered images).
    \item To the best of our knowledge, we are the first to benchmark open-source mLLM families, not only on Table QA and Attribution performances, but also under confidence alignment and uncertainty calibration.
    \item Our findings reveal that mLLMs often struggle in spatial QA tasks compared to spatial attribution in a text-in-vision paradigm. 
\end{itemize}

\begin{figure}[htbp]
    \centering
    \includegraphics[width=0.80\textwidth]{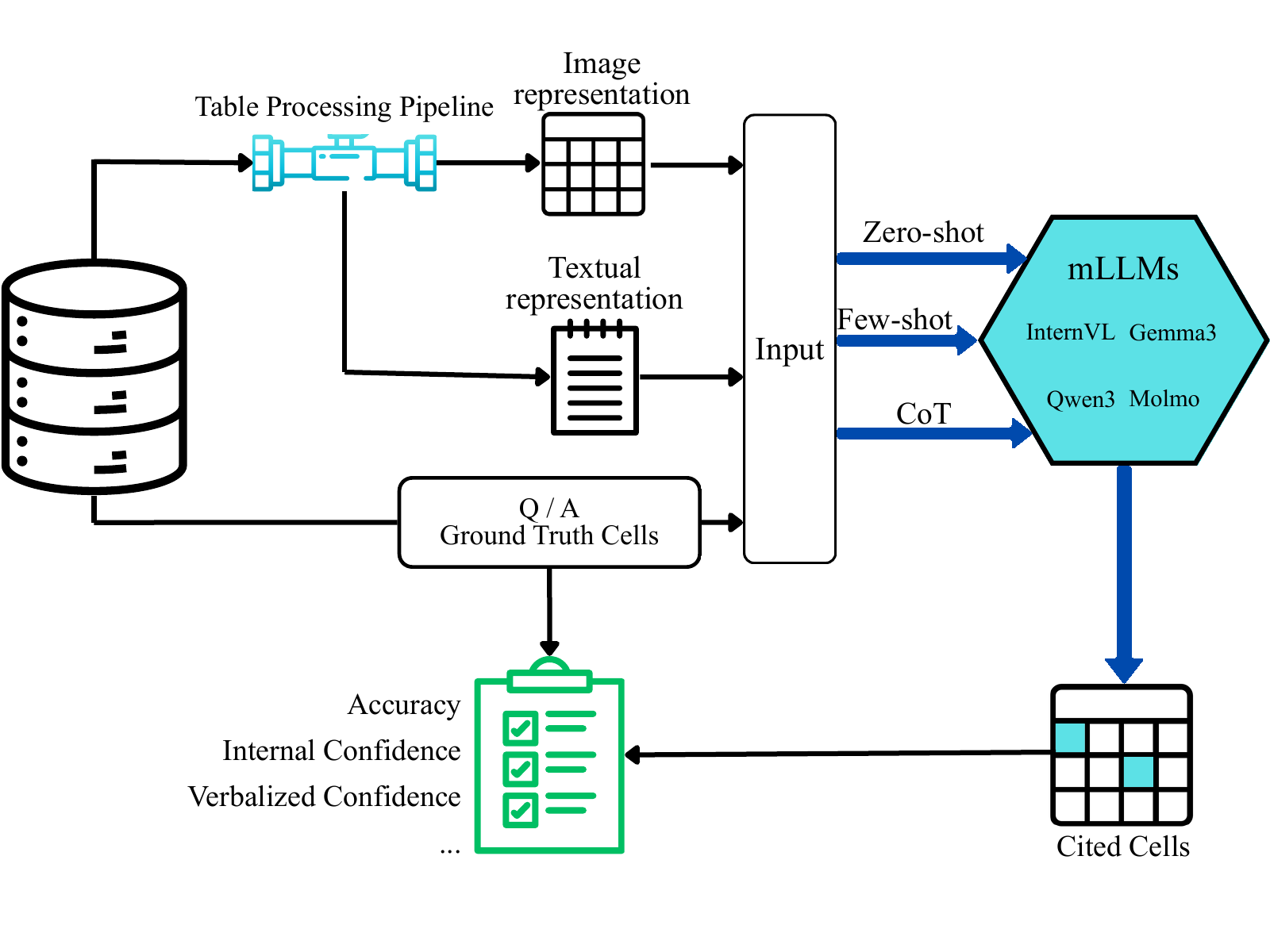}
    \vspace{-1em}
    \caption{A brief overview of the general workflow of our proposed framework: ViTaB-A benchmarking.}
    \label{fig:flow}
    \vspace{-2em}
\end{figure}

\section{Related Works}
\vspace{-0.5em}
\textbf{Question Answering over Structured Data:} Early neural methods such as TaPas~\citep{herzig2020tapas} and structure-aware transformers~\citep{zhang2020table} showed table reasoning without usng hand-crafted query programs. Datasets like HiTab~\citep{cheng-etal-2022-hitab} and HiBench~\citep{jiang2025hibench} introduce hierarchical and structured challenges. With LLMs and mLLMs, several studies report moderate QA performance on both text and image table formats~\citep{fang2024large, deng-etal-2024-tables, zheng-etal-2024-multimodal}.

\textbf{Attribution and Grounding in LLMs:} Work on hallucination and grounding shows that models often produce fluent but unsupported claims~\citep{huang2025survey}. Prompted citation generation still yields incorrect or unverifiable references~\citep{gao2023enabling}. Benchmarks and systems for attributed QA show evidence selection is hard in practice~\citep{bohnet2022attributed, vankov2025consens, radevski2025synthesizing}. Recent efforts target structured table attribution directly, e.g., MATSA and TabCite~\citep{mathur2024matsa} and automatic attribution benchmarks~\citep{hu2025can}.

\textbf{Optimization Objectives and Confidence:} Modern LLMs use next-token pretraining followed by instruction tuning and RL fine-tuning that reward helpful answers~\cite{ouyang2022training}. Evaluation suites emphasize final-answer correctness~\citep{liang2022holistic}. Systems such as WebGPT and multi-agent pipelines add components to improve grounding~\citep{nakano2021webgpt, mathur2024matsa}. Research on confidence and calibration reveals frequent misalignment between confidence and accuracy~\citep{kumar-etal-2024-confidence, geng-etal-2024-confsurvey, ye2024benchmarkingllmsuncertaintyquantification, groot-valdenegro-toro-2024-overconfidence}.

\textbf{Representation Effects:} Representation and layout affect model behavior. Comparisons of text vs image table inputs and multimodal spatial challenges are reported in prior work~\citep{deng-etal-2024-tables, zheng-etal-2024-multimodal, wang-etal-2024-docllm, liu2025spatialreasoningmultimodallarge}.

\section{Methodology}
\vspace{-0.5em}

As prior works have demonstrated, mLLMs can produce correct answers, even with spurious calculations or hallucinated reasoning. We conduct a comprehensive study on visual table attribution to investigate trust and reliability in mLLMs. We focus on two major research questions:
\vspace{-0.25em}
\begin{enumerate}\setlength{\itemsep}{2pt}\setlength{\parskip}{0pt}\setlength{\topsep}{2pt}
    \item \textit{How accurately do mLLMs identify table cells that support a given answer?}
    \item \textit{Does a model's confidence score reliably reflect the correctness of its attribution?}
\end{enumerate}

\vspace{-0.25em}
To address these questions, we benchmark mLLM attribution accuracy and uncertainty behavior across model family, input representations, and prompting strategies. \autoref{fig:flow} details the workflow of our benchmarking approach.

\subsection{Task Formulation}
\label{sec:task_form}

We analyze the \emph{spatial intelligence} of an mLLM-family using structured tabular data attribution. 
\ignorespacesafterend
Tables provide a controlled grounding substrate: evidence is discrete and compositional, and can be referenced unambiguously via row-column coordinates. This makes attribution a simple but revealing proxy for spatial competence: a model must align language with precise table structure (cells / rows / columns) rather than merely produce plausible text~\citep{liu2025spatialreasoningmultimodallarge}.

Each instance consists of a natural-language question $q$, a provided (correct) answer $a$, and a table $T$. The answer is \emph{given} to the model to shift the burden from generation to grounding: the model is asked to identify \emph{where} the support for $a$ resides in $T$. This isolates spatial grounding ability and enables controlled comparisons across model variants.
Formally, let $\mathcal{I}(T)=\{(i,j)\mid i\in[m],\,j\in[n]\}$ denote the set of cell indices for an $m\times n$ table. Given $(q,a,T)$, the model outputs a set of cited cells
\[
\hat{S} = f_{\theta}(q,a,X_r)\subseteq \mathcal{I}(T),
\]
returned as row--column indices (including header cells when they are part of the evidence)~\citep{mathur2024matsa}. Here $X_r$ represents the table encoding.

Finally, to generalize across model scale and inference techniques, we assess the aforementioned properties on multiple model families with varying training parameters under standard in-context learning paradigms: zero-shot~\citep{brown2020language}, few-shot~\citep{brown2020language}, and CoT~\citep{wei2022chain} prompting. All prompts used in this paper are provided in Appendix~\ref{app:full_prompts}.

\begin{tcolorbox}[verticalbox=RoyalBlue, title=\textbf{Vertical 1: Representation Gap (Visual vs.\ Text)}]
We vary the table representation
\[
r\in \mathcal{R}\triangleq\{\textsf{image},\textsf{markdown},\textsf{json}\}, \qquad X_r=\mathrm{Enc}_r(T),
\]
where $X_{\textsf{image}}$ is a rendered table image and $X_{\textsf{markdown}},X_{\textsf{json}}$ are structured text encodings.
This contrasts attribution under true visual parsing vs.\ attribution under serialization, exposing whether grounding is genuinely multimodal~\citep{deng-etal-2024-tables}.
For $\textsf{image}$ tables, we apply semantics-preserving perturbations
\[
X'_{\textsf{image}}=\pi(X_{\textsf{image}}),
\]
where $\pi$ changes appearance without changing cell content or layout (e.g., header/cell color changes, font/style changes). This allows us to assess the aggregate impact of superficial stylistic variations on attribution quality of rendered image data.
\end{tcolorbox}

\begin{tcolorbox}[verticalbox=ForestGreen, title=\textbf{Vertical 2: Reliability and Confidence Alignment}]
Beyond \emph{what} cells are cited, we assess whether the model can \emph{reliably communicate} attribution correctness. 
We compare (i) \emph{internal confidence} derived from token-level likelihoods of the citation output with (ii) \emph{verbatim (verbalized) confidence} elicited as an explicit self-report; misalignment between the two is a known failure mode~\citep{geng-etal-2024-confsurvey, kumar-etal-2024-confidence}.
\end{tcolorbox}

\subsection{Benchmark}
\label{sec:benchmark}

Our experiments are conducted on \textbf{ViTaB-A}, which is constructed using the HiTab~\citep{cheng-etal-2022-hitab} dataset, which contains question-answer pairs grounded in structured tables with annotated evidence cells. HiTab provides which ground truth reference table entries are required to support a correct answer, making it well-suited for attribution-centric evaluation.

We standardize attribution across representations by augmenting each table with explicit row and column labels. This enables unambiguous cell references (e.g., B3, E7) across all experimental conditions for accurate evaluation.

We present the tables to models using three different representations: \textbf{(1) JSON; (2) Markdown; (3) Rendered images}. Such setup allows us to study the attribution behavior under both structured textual and visual inputs. For image-based tables, we additionally introduce controlled visual perturbations that preserve the underlying tabular content while altering table appearance. These perturbations include variations in header color (red, blue, and green) and font style (Arial and Times New Roman). For each representation, we select 200 tables as the visual table attribution benchmark.

\subsection{Model Setup}
\label{sec:model_setup}

We evaluate a diverse set of mLLMs with varying architectures and parameter scales to investigate the capabilities of visual table attribution across model families. Specifically, we consider the \textbf{Gemma-3} family (4B, 12B, and 27B), \textbf{InternVL3.5} models (4B, 8B, 14B, and 38B), \textbf{Qwen3-VL} vision-language models (2B, 4B, 8B, and 32B), and the \textbf{Molmo2} family (4B and 8B) to assess evolution of attribution accuracy and uncertainty across model scale within a family as well as architectural differences across families.

For clarity and standardized comparison, we focus our discussion in Section~\ref{sec:results} on the \textbf{4B-scale models} from each family. However, a comprehensive analysis covering all evaluated model sizes and configurations is provided in Appendix~\ref{app:full_results}.

\subsection{Attribution Metrics}
\label{sec:att_metrics}
\vspace{-0.5em}
We evaluate metrics that capture both the statistical accuracy of attribution and the alignment between model's internal and expressed confidence in attribution.

\vspace{-0.5em}
\subsubsection{Statistical Accuracy}
\vspace{-0.5em}
We extract the row–column indices referenced by the model and compare them against the ground-truth attribution set. 
We compute cell-level accuracy measure to see how accurately the model returns the correct evidence cells. We also report row-wise and column-wise accuracy scores to gain insight into the model's localization ability. These metrics help understand if the inaccuracy stems from localization errors or failure to pinpoint exact cells. Additionally, we observe cell-wise, row-wise and column-wise precision, recall, and F1 scores which are included in the Appendix~\ref{app:full_results}.

Collectively, these metrics provide quantitative evidence of current mLLMs’ ability to accurately retrieve and localize attribution references.

\vspace{-0.5em}
\subsubsection{Confidence-Accuracy Alignment}
\vspace{-0.5em}
\label{sec:conf_align}
Confidence-Probability Alignment~\citep{kumar2024confidencehoodinvestigationconfidenceprobability} refers to the correlation between a model's internal confidence and the verbalized certainty. We derive \textit{Internal Confidence} directly from the model's answer level probability and reflects how strongly the model internally prioritizes a selected attribution over the rest. On the otherhand, the \textit{Verbalized Certainty} is defined as the model's explicit expression of its confidence level through the evaluation of its natural language answer. High correlation between these two metrics corroborate the transparency and reliability of the model for our attribution task.

\vspace{-0.5em}
\paragraph{Internal Confidence:} In visual table attribution, we define internal confidence as the normalized probability of a predicted cell relative to all candidate cells. For each output token $\mathcal{T}_i$, we can convert the logits $L(\mathcal{T}_i)$ to probability using the softmax function.
\begin{equation*}
    P_t(\mathcal{T}_i) = \frac{e^{L(\mathcal{T}_i)}}{\sum_j e^{L(\mathcal{T}_j)}}.
\end{equation*}

Each candidate cell $c$ may correspond to multiple tokenizations. Let $\mathcal{T}^C$ denote the set of token IDs associated with a cell $c$. We define the raw cell confidence as the geometric mean probability among the corresponding tokens.
\begin{equation*}
    P(c) = \left( \prod_{t \in \mathcal{T}^C} P_t \right)^{\frac{1}{|\mathcal{T}^C|}}
\end{equation*}

We normalize the raw $P(c)$ scores to obtain the adjusted internal confidence,
\begin{equation*}
    P_{IC}(c) = \frac{P(c)}{\sum_{c' \in \mathcal{C}} P(c')},
\end{equation*}
where $\mathcal{C}$ denotes the set of all table cells. For answers involving multiple cell citations, we aggregate the individual $P_{IC}(c)$ scores with pooling functions (e.g. mean, max, or product) to obtain a single confidence score.
Higher $P_{IC}(c)$ signifies greater model confidence in that output cell

\vspace{-0.5em}
\paragraph{Verbalized Certainty:} Verbalized certainty is the model's evaluation of explicit confidence level in its own natural language answer. Inspired by~\citep{kumar2024confidencehoodinvestigationconfidenceprobability}, we develop a Confidence Querying Prompt (CQP) that asks the model to analyze the expressed certainty in the context of the question, answer, predicted cells, and the table representation i.e. all possible candidate cells. 

The model selects one of six ranked certainty levels: \textit{Very Certain}, \textit{Fairly Certain}, \textit{Moderately Certain}, \textit{Somewhat Certain}, \textit{Not Certain}, and \textit{Very Uncertain}. These ordinal levels are mapped to confidence scores in the interval $[0,1]$ with increments of $0.2$, enabling quantitative comparison with internal confidence estimates.

The query effectively prompts the model to adopt an observational perspective and analyze the certainty of its answer. Additionally, by providing all cell options allows the model to contextualize its chosen response that leads to more informed confidence judgments and further implicit verification than only isolated evaluations.

The complete CQP formulation is provided in Appendix~\ref{app:cqp_prompt}.

\vspace{-0.5em}
\paragraph{Alignment Evaluation:} To quantify confidence-accuracy alignment, we compute how well the confidence scores--internal or verbalized--correspond to the attribution accuracy using \textit{Brier Score}~\citep{glenn1950verification}. Brier score directly penalizes the discrepancy between individual confidence estimate and attribution accuracy, providing a clear insight into model misalignment.

For each model response $i$, the Brier Score is calculated as follows:
\vspace{-0.5em}
\begin{equation*}
\text{P} = \frac{1}{n} \sum_{i=1}^{n} (c_i - a_i)^2.
\end{equation*}
where $c_i \in [0,1]$ is the confidence score and $a_i \in [0, 1]$ is the accuracy measure. We define the alignment score as,
\begin{equation*}
\mathcal{A} = 1 - P.
\end{equation*}
where a higher alignment score reflects better calibration between expressed (or internal) belief and actual prediction quality.

Additionally, we perform uncertainty quantification as presented by~\cite{ye2024benchmarkingllmsuncertaintyquantification} for rigorous and model-agnostic uncertainty estimates, the detailed experiments and results of which are reported in Appendix~\ref{sec:uncert_quant}.

\section{Results}
\label{sec:results}
\subsection{Subpar Attribution Despite Reasonable Question Answering}
\label{sec:attribution_results}

Our findings show that mLLM's are not inherently bad at question answering (QA) over structured data, but their performance decreases significantly for attribution tasks. As shown in \autoref{tab:accuracy_results}, QA accuracy remains relatively stable across models and modalities, typically around 50-60\%. In contrast, attribution accuracy reported in \autoref{tab:accuracy_results} is dramatically lower, ranging from near-random performance in JSON to around 33\% in images.

\begin{table*}[!htbp]
\small
\centering
\setlength{\aboverulesep}{0pt}
\setlength{\belowrulesep}{0pt}
\caption{Model Accuracy in QA vs Attribution in \% Across Prompting Strategies Across Open-source Models (for 4B Parameter); \textbf{Note}: \textcolor{ForestGreen}{green} depicts overall best model, and \textcolor{red}{red} depicts worst.}
\label{tab:accuracy_results}
\begin{tabular}{@{}llcccccccc@{}}
\toprule \hline
\multirow{2}{*}{\textbf{Strategy}} & \multirow{2}{*}{\textbf{Model}} 
& \multicolumn{2}{c}{\textbf{Markdown}} 
& \multicolumn{2}{c}{\textbf{JSON}} 
& \multicolumn{2}{c}{\textbf{Images}} 
& \multicolumn{2}{c}{\textbf{Average}} \\
\cmidrule(lr){3-4} \cmidrule(lr){5-6} \cmidrule(lr){7-8} \cmidrule(lr){9-10}
& & QA & Attr. & QA & Attr. & QA & Attr. & QA & Attr. \\
\midrule
\multirow{4}{*}{Zero Shot}
& Qwen3-VL        & 60.00 & 35.50 & 61.50 & 01.00 & 62.00 & 45.40 & 61.16 & 27.30 \\
& Gemma3       & 43.00 & 13.40 & 40.00 & 00.80 & 28.00 & 16.60 & \textcolor{red}{37.00} & \textcolor{red}{10.27} \\
& Molmo2         & 55.50 & 19.50 & 59.50 & 01.00 & 48.50 & 33.60 & 54.50 & 18.03 \\
& InternVL3.5  & 63.50 & 42.50 & 64.00 & 01.50 & 60.50 & 53.10 & \textcolor{ForestGreen}{62.66} & \textcolor{ForestGreen}{32.37} \\
\midrule
\multirow{4}{*}{Few Shot}
& Qwen3-VL        & 62.00 & 21.50 & 61.50 & 01.50 & 60.50 & 34.90 & 61.33 & 19.30 \\
& Gemma3       & 39.50 & 6.20  & 36.00 & 00.60 & 25.00 & 14.20 & \textcolor{red}{34.00} & \textcolor{red}{07.00} \\
& Molmo2         & 58.00 & 18.00 & 56.50 & 00.50 & 45.00 & 20.20 & 53.00 & 12.90 \\
& InternVL3.5  & 64.00 & 36.50 & 64.00 & 01.00 & 58.50 & 52.60 & \textcolor{ForestGreen}{62.16} & \textcolor{ForestGreen}{30.03} \\
\midrule
\multirow{4}{*}{CoT}
& Qwen3-VL        & 61.00 & 49.00 & 61.00 & 01.00 & 59.50 & 44.40 & \textcolor{ForestGreen}{60.33} & \textcolor{ForestGreen}{31.47} \\
& Gemma3       & 41.00 & 10.00 & 38.00 & 00.20 & 27.50 & 14.10 & \textcolor{red}{35.66} & \textcolor{red}{08.10} \\
& Molmo2         & 55.50 & 15.50 & 57.50 & 01.00 & 50.00 & 22.90 & 54.33 & 13.13 \\
& InternVL3.5  & 59.00 & 39.00 & 62.50 & 00.50 & 59.00 & 54.70 & \textcolor{ForestGreen}{60.33} & 31.40 \\
\midrule
\multicolumn{2}{l}{\textbf{Average}} 
& 55.08 & 25.55 
& \textbf{55.16} & 00.88 
& 48.66 & \textbf{33.89} 
& -- & -- \\
\bottomrule \hline
\end{tabular}
\end{table*}

This massive drop shows that poor attribution performance cannot be explained by weak reasoning or answer generation. Instead, models often identify the correct answer but fail to reliably point to the specific rows and columns that support it. Answer correctness and attribution quality therefore appear to be separate capabilities. As a result, models may appear reliable based on QA benchmarks while remaining unsuitable for applications that require traceability or auditability such as those in regulated industries like banking, healthcare and law. We discuss potential causes of this disconnect, including training objectives and evaluation practices in Section \ref{sec:discussion}.

\subsection{Attribution is Easier in Images than in Text}

From \autoref{tab:accuracy_results}, we see that attribution performance varies substantially across input modalities. Models perform best when tables are presented as images, followed by Markdown, with JSON being by far the most difficult format. Average attribution accuracy on JSON is below 1\%, compared to over 30\% for images.

One likely explanation is that images preserve the spatial and visual hierarchy of tables, allowing models to rely on layout-based cues such as row alignment, column boundaries, and visual grouping. Prior work has shown that multimodal models can effectively leverage spatial structure in document images for tasks such as table understanding and information extraction \citep{zheng-etal-2024-multimodal, wang-etal-2024-docllm}. By contrast, textual formats like JSON lack visual structure and encode hierarchy only through nested text, requiring models to try and understand structure from sequences - a setting that has been shown to be difficult both theoretically and empirically \citep{jiang2025hibench, hahn2020theoretical}.

Interestingly, this trend reverses for QA accuracy. As shown in \autoref{tab:accuracy_results}, models often perform better at answering questions in textual formats than in images, despite performing worse at attribution in those same formats. One plausible explanation is that QA places weaker grounding requirements than attribution: models can often infer the correct answer from partial cues or broadly relevant context without needing to explicitly identify the supporting evidence \citep{bohnet2022attributed, radevski2025synthesizing}. Prior work has also suggested that reliably assessing attribution and context grounding is significantly more challenging than answer generation itself \citep{hu2025can, vankov2025consens}. Together, these results suggest that while textual formats are often sufficient for producing correct answers, they pose a significantly greater challenge for precise and reliable attribution.

Overall, models attribute most reliably in image-based tables and struggle in textual formats, particularly JSON. While textual formats support accurate answer generation, they make citation significantly harder. This supports our hypothesis from earlier that question answering and attribution/citation are two distinct tasks.

\subsection{Models are Better at Citing Rows than Columns}

Across modalities (except JSON) and prompting strategies, models are substantially better at identifying the correct row rather than the correct column (\autoref{tab:row_col_attribution_accuracy}). Averaged across models and prompts, row accuracy is approximately 1.3-2$\times$ higher than column accuracy for Markdown and image-based tables.

\begin{table}[t]
\centering
\setlength{\aboverulesep}{0pt}
\setlength{\belowrulesep}{0pt}
\small
\caption{Row vs Column Accuracy in \% Across Modalities and Prompting Strategies, Across Open-source Models (for 4B Parameter); \textbf{Note}: \textcolor{ForestGreen}{green} depicts overall best model, and \textcolor{red}{red} depicts worst.}
\label{tab:row_col_attribution_accuracy}
\begin{tabular}{@{}llcccccc@{}}
\toprule
& & \multicolumn{2}{c}{\textbf{Markdown}} & \multicolumn{2}{c}{\textbf{JSON}} & \multicolumn{2}{c}{\textbf{Images}} \\
\cmidrule(lr){3-4} \cmidrule(lr){5-6} \cmidrule(lr){7-8}
\textbf{Strategy} & \textbf{Model} & \textbf{Row} & \textbf{Column} & \textbf{Row} & \textbf{Column} & \textbf{Row} & \textbf{Column} \\
\midrule
\multirow{4}{*}{Zero Shot} 
& Qwen3-VL & 73.75 & 48.83 & 10.13 & 36.92 & 78.01 & 58.55 \\
& Gemma3 & 70.86 & 23.17 & 5.83 & 21.08 & 47.82 & 30.98 \\
& Molmo2 & 68.75 & 27.58 & 9.16 & 23.33 & 58.51 & 56.13 \\
& InternVL3.5 & 78.25 & 36 & 4.04 & 42.33 & 69.24 & 59.79 \\
\midrule
\multirow{4}{*}{Few Shot} 
& Qwen3-VL & 77 & 29.83 & 10.53 & 11.75 & 79.05 & 46.65 \\
& Gemma3 & 55.37 & 16.33 & 1.83 & 11.83 & 42.51 & 31.28 \\
& Molmo2 & 72.5 & 24.08 & 9.3 & 16.33 & 40.96 & 35.16 \\
& InternVL3.5 & 64.8 & 18.08 & 3.08 & 18.25 & 59.43 & 16.43 \\
\midrule
\multirow{4}{*}{CoT} 
& Qwen3-VL & 82.5 & 59.17 & 5.83 & 42.92 & 77.32 & 57.59 \\
& Gemma3 & 64.79 & 17.83 & 3.38 & 21.25 & 46.13 & 35.56 \\
& Molmo2 & 68.8 & 18 & 4.88 & 14 & 54.73 & 37.55 \\
& InternVL3.5 & 82.25 & 41.5 & 3.24 & 53.83 & 78.77 & 64.82 \\
\midrule
\multicolumn{2}{l}{\textbf{Average}} & \textcolor{ForestGreen}{\textbf{71.63}} & \textcolor{red}{30.03} & \textcolor{red}{5.93} & \textcolor{ForestGreen}{\textbf{26.15}} & \textcolor{ForestGreen}{\textbf{61.04}} & \textcolor{red}{44.2} \\
\bottomrule
\end{tabular}
\end{table}

One possible reason for this disparity is that rows and columns play very different roles in a table. Rows often represent complete, meaningful records, such as a single person, product, or transaction. Columns, on the other hand, represent abstract attributes or fields, such as dates, categories, or numerical properties. Prior work on table reasoning has shown that identifying and reasoning about the correct column -- often referred to as schema linking -- is especially difficult for language models, particularly when column headers are ambiguous or require implicit interpretation \citep{zhang2020table, herzig2020tapas}. In contrast, rows are easier to localize because they more closely match how entities and examples are described in natural language.

This consistent disparity suggests that fine-grained attribution -- especially identifying the correct field within a record -- remains a major unresolved challenge for mLLMs. 

\begin{table}[!htbp]
\centering
\small
\setlength{\aboverulesep}{0pt}
\caption{Confidence-Accuracy correlation for Internal and Verbal; Across Multiple Modalities.}
\label{tab:acc_internal_verbal}
\vspace{-0.8em}
\begin{tabular}{@{}llcccccc@{}}
\toprule
& & \multicolumn{2}{c}{\textbf{Markdown}} & \multicolumn{2}{c}{\textbf{JSON}} & \multicolumn{2}{c}{\textbf{Images}} \\
\cmidrule(lr){3-4} \cmidrule(lr){5-6} \cmidrule(lr){7-8}
\textbf{Strategy} & \textbf{Model} & \textbf{Internal} & \textbf{Verbal} & \textbf{Internal} & \textbf{Verbal} & \textbf{Internal} & \textbf{Verbal} \\
\midrule
\multirow{4}{*}{Zero Shot} 
& Qwen3-VL & 0.56 & 0.69 & 0.42 & 0.62 & 0.60 & 0.62 \\
& Gemma3 & 0.45 & 0.40 & 0.27 & 0.32 & 0.41 & 0.39 \\
& Molmo2 & 0.73 & 0.56 & 0.82 & 0.68 & 0.83 & 0.38 \\
& InternVL3.5 & 0.64 & 0.65 & 0.74 & 0.83 & 0.77 & 0.67 \\
\midrule
\multirow{4}{*}{Few Shot} 
& Qwen3-VL & 0.52 & 0.73 & 0.59 & 0.80 & 0.55 & 0.55 \\
& Gemma3 & 0.27 & 0.26 & 0.58 & 0.16 & 0.40 & 0.38 \\
& Molmo2 & 0.76 & 0.62 & 0.84 & 0.81 & 0.77 & 0.27 \\
& InternVL3.5 & 0.55 & 0.72 & 0.79 & 0.83 & 0.53 & 0.71 \\
\midrule
\multirow{4}{*}{CoT} 
& Qwen3-VL & 0.65 & 0.69 & 0.36 & 0.63 & 0.61 & 0.59 \\
& Gemma3 & 0.27 & 0.36 & 0.40 & 0.29 & 0.36 & 0.38 \\
& Molmo2 & 0.68 & 0.62 & 0.88 & 0.65 & 0.77 & 0.33 \\
& InternVL3.5 & 0.58 & 0.69 & 0.51 & 0.85 & 0.65 & 0.73 \\
\midrule
\multicolumn{2}{l}{\textbf{Average}} 
& 0.555 & 0.583 & 0.601 & 0.620 & 0.600 & 0.500 \\
\bottomrule
\end{tabular}
\end{table}

\subsection{Lack of Statistically Significant Alignment between Confidence and Attribution Accuracy}
\label{sec:conf_acc_alignment}

\begin{figure}[htbp]
    \centering    \includegraphics[width=0.8\textwidth]{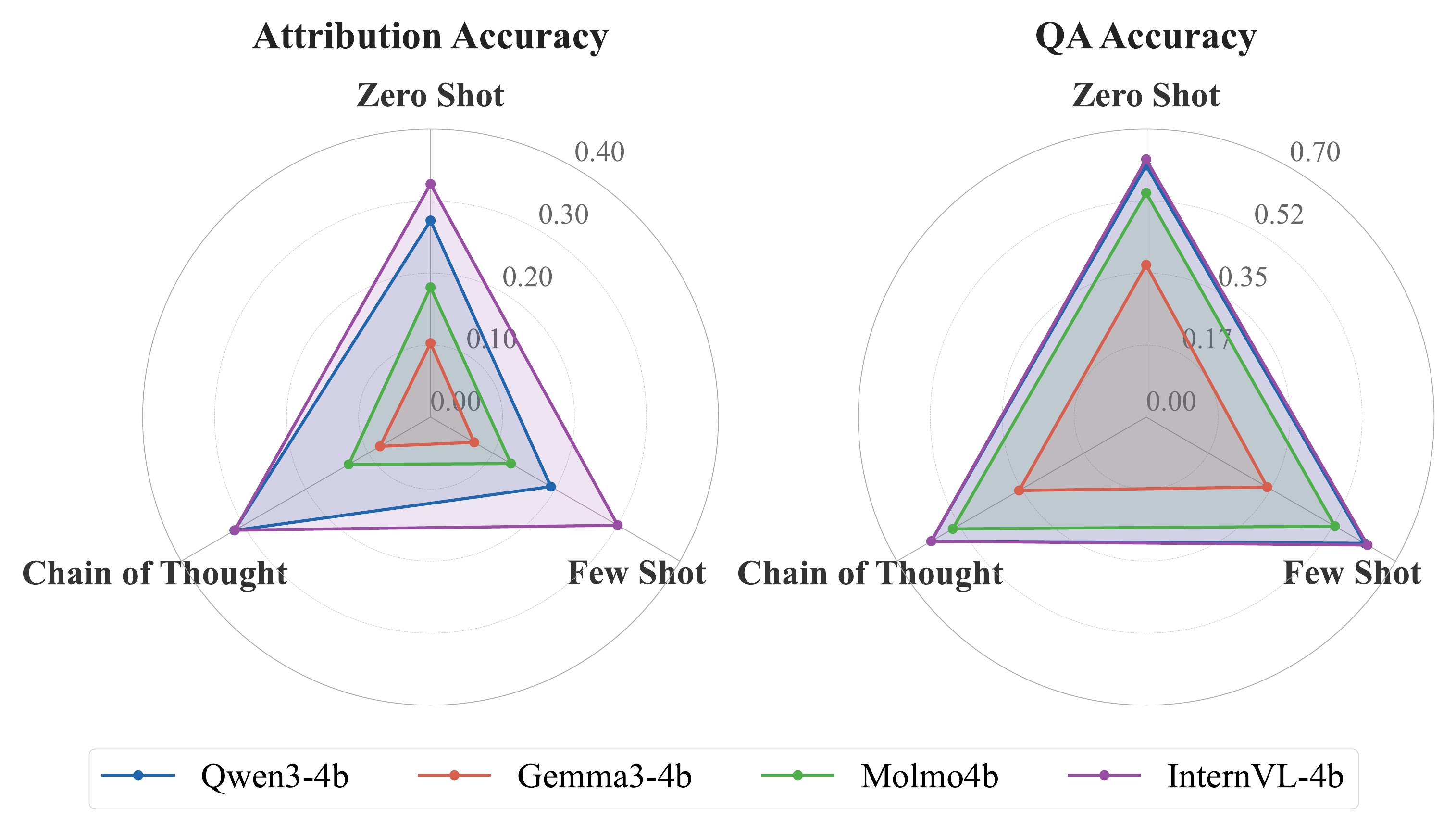}
    \caption{Radar charts comparing model families across attribution accuracy, QA accuracy, and confidence gap ($1 - |\text{verbal} - \text{internal}|$) under different prompting strategies.}
    \label{fig:model_family_radar}
\end{figure}

There is no clear advantage of using confidence as an indicator for attribution ability. From \autoref{tab:acc_internal_verbal}, we observe that the confidence-accuracy alignment scores for the attribution task vary across models, representations, and prompting paradigms and display no consistent or strong correlation even though the confidence scores generally lie between 60-80\% (Appendix \autoref{tab:confidence_scores}). Across all models (except Molmo2), internal confidence and accuracy alignment is $< 70\%$. And even for Molmo2-, which exhibits high internal alignment for textual representation, attribution accuracy ranks 3rd among other model families (\autoref{fig:model_family_radar}). Verbal alignment shows similar subpar scores, establishing the unreliability of confidence scores in attribution quality comparison. 

The observation is consistent with prior research on confidence. \cite{groot-valdenegro-toro-2024-overconfidence, kumar-etal-2024-confidence} highlight the disparity between verbalized and underlying token-level confidence scores and ~\cite{geng-etal-2024-confsurvey} reports that token probabilities are not inherently well-aligned with task accuracies. Furthermore, \cite{ye2024benchmarkingllmsuncertaintyquantification} show that naive confidence measures alone are insufficient metrics and require post-processing to meaningfully reflect reliability. 

Collectively, these outcomes support our conclusion that confidence measures should not be viewed as a reliable indicator for attribution quality.

\autoref{fig:model_family_radar} compares model families across attribution accuracy, QA accuracy, and confidence gap under zero-shot, few-shot, and chain-of-thought prompting. Clear differences emerge across model families, showing that attribution performance depends heavily on the underlying model.

\subsection{Summary}

Overall, our results show that current mLLMs struggle to reliably attribute or cite information in structured data. Across models and prompting strategies, citation accuracy is consistently low, particularly in structured formats such as JSON. At the same time, model confidence remains moderate to high, even when attribution performance collapses. This indicates that confidence cannot be reliably used to estimate citation correctness. 

Importantly, this means that adding a second step that verifies citations using confidence may not reduce user risk. Confidence appears more aligned with answer generation than with attribution fidelity. As a result, models may appear reliable wile still failing to provide accurate traceability. 

Attribution performance further depends strongly on how structured data is represented. Models attribute most reliably when tables are presented as images and perform substantially worse on textual formats such as Markdown and JSON, with JSON being the most challenging. Across all modalities, models are also significantly better at identifying the correct row than the correct column, highlighting continued difficulties with fine-grained localization. Finally, we observe clear difference across model families. InternVL3.5-4b achieves the strongest attribution and QA accuracy.

\section{Conclusion \& Future Works}
\label{sec:discussion}

The gap between question answering and attribution quality reflects how current mLLMs are trained and evaluated. Most instruction-tuning and alignment pipelines optimize models to produce correct and helpful answers, but do not explicitly reward precise citation or faithful attribution to specific data fields \citep{ouyang2022training}. As a result, models can often answer questions correctly without reliably identifying the rows and columns that support those answers. Prior work such as WebGPT shows that accurate citation requires task-specific objectives, rather than emerging naturally from standard training pipelines \citep{nakano2021webgpt}. Our proposed benchmark, ViTaB-A facilitates in affirming that QA accuracy and attribution accuracy are separate capabilities, and progress on one does not guarantee progress on the other.

This separation is reinforced by current evaluation practices. Benchmarks such as HELM focus primarily on answer accuracy, robustness, and calibration, while structured attribution and traceability receive little attention \citep{liang2022holistic}. Although prompting strategies can slightly affect attribution, they do not close the gap, suggesting that inference-time methods alone are insufficient. This creates a feedback loop where models are optimized and compared mainly on QA accuracy, even though attribution remains unreliable.

These findings point to several directions for future work. Attribution should be treated as a first-class objective, with training signals that directly optimize row and column localization. Finally, the high verbal confidence models express despite incorrect attribution raises concerns for user trust, especially in high-stakes domains. Overall, our results suggest that improving QA accuracy alone is not sufficient, and that reliable structured attribution requires dedicated research attention.

\section*{Acknowledgments}
We thank Archit Singh and Arjun Khetan for help during the early stages of this project while it was conducted as a class project; their contributions to initital experiments were valuable. We are also grateful to Bhanu Tokas, Aman Panwar, Pranav K. Nayak and, all the other anonymous reviewers who read early drafts and offered constructive feedback that improved the paper.

\bibliography{iclr2026_conference}
\bibliographystyle{iclr2026_conference}

\appendix
\section{Appendix}

\subsection{Attribution Task Prompts}
\label{app:full_prompts}
\paragraph{Zero-shot Technique:}
\begin{quote}
\footnotesize
\begin{Verbatim}[commandchars=\\\{\}]
You are a table analysis expert. Your task is to identify which 
cell(s) in the table contain or support the given answer to the 
question.

TABLE: \{table\}

QUESTION: \{question\}
ANSWER: \{answer\}

TASK: Identify the cell coordinate(s) that contain or directly 
support this answer. Use Excel-style coordinates where columns 
are letters (A, B, C, ...) and rows are numbers (1, 2, 3, ...).

RESPONSE FORMAT: Return ONLY the cell coordinates in Excel formula 
format. 
Examples:
- Single cell: "=E7" or "=B3"
- Multiple cells: "=A2" or list them separately: "=A2, =B2, =C2"
- If the answer involves a formula (sum, average, etc.), 
you may use: "SUM(C3:C10)" or "=A1+B2"

IMPORTANT: Do NOT repeat the question, table, or instructions. 
Output ONLY the cell coordinates.

ATTRIBUTED CELLS:
\end{Verbatim}
\end{quote}

\paragraph{Few-shot Technique:}
\begin{quote}
\footnotesize
\begin{Verbatim}[commandchars=\\\{\}]
You are a table analysis expert. Your task is to identify which 
cell(s) in the table contain or support the given answer to the 
question.

Here is an example:

EXAMPLE:
TABLE:
\{example1_table\}
QUESTION: \{example1_question\}
ANSWER: \{example1_answer\}
ATTRIBUTED CELLS: \{example1_cells\}

Now analyze this table:

TABLE: \{table\}

QUESTION: \{question\}
ANSWER: \{answer\}

IMPORTANT: Do NOT repeat the example, question, table, or 
instructions. Output ONLY the cell coordinates in formula 
format.

ATTRIBUTED CELLS:
\end{Verbatim}
\end{quote}

\paragraph{Chain-of-Thought Technique:}
\begin{quote}
\footnotesize
\begin{Verbatim}[commandchars=\\\{\}]
You are a table analysis expert. Your task is to identify which 
cell(s) in the table contain or support the given answer to the 
question.

TABLE: \{table\}

QUESTION: \{question\}
ANSWER: \{answer\}

Let's think step by step:

1. First, understand what the question is asking for.
2. Then, locate where the answer "\{answer\}" appears or can be 
derived from in the table.
3. Identify the specific cell coordinate(s) using Excel-style 
notation (columns as letters A, B, C... and rows as numbers 
1, 2, 3...).
4. If the answer is computed from multiple cells (e.g., a sum), 
express it as a formula like "SUM(C3:C10)" or "=A1+B2".
5. For simple cell references, use the format "=E7" or "=B3".

IMPORTANT: Do NOT repeat the question or table in your reasoning.

REASONING:

Based on the above reasoning, provide ONLY the final cell coordinates 
in Excel formula format (e.g., "=E7", "SUM(C3:C10)", or "=A1+B2"). 
Do NOT repeat your reasoning or the question.

ATTRIBUTED CELLS:
\end{Verbatim}
\end{quote}

\paragraph{Confidence Querying Prompt (CQP) :}
\label{app:cqp_prompt}
\begin{quote}
\footnotesize
\begin{Verbatim}[commandchars=\\\{\}]
A language model was analyzing a table to identify cells that 
support an answer.
\textcolor{red}{TABLE: \{table\}}
\textcolor{red}{QUESTION: \{question\}}
\textcolor{Green}{ANSWER: \{answer\}}
\textcolor{Green}{All possible cell coordinates in this table: \{all_cells_list\}}
\textcolor{Green}{The model selected the following cell(s): \{predicted_cells\}}

Critically analyze whether the identified cells correctly support 
the answer given all the available cells. Check for missed cells 
or inclusion of irrelevant cells.
How certain are you that the model's cell selection is completely correct?

\textcolor{blue}{a. Very Certain (No doubts at all)}
\textcolor{blue}{b. Fairly Certain (Minor doubts)}
\textcolor{blue}{c. Moderately Certain (Some doubts)}
\textcolor{blue}{d. Somewhat Certain (Significant doubts)}
\textcolor{blue}{e. Not Certain (Likely incorrect)}
\textcolor{blue}{f. Very Uncertain (Definitely incorrect)}

Answer with just the letter (a-f):
\end{Verbatim}
\end{quote}

\subsection{Confidence Scores:}
\begin{table}[htb]
\centering
\small
\caption{Internal and Verbal Confidence Scores in \% Across Modalities and Prompting Strategies. Difference is calculated as \texttt{$|$avg(verbal) - avg(internal)$|$} for each row.}
\label{tab:confidence_scores}
\begin{tabular}{@{}llccccccc@{}}
\toprule
& & \multicolumn{2}{c}{\textbf{Markdown}} & \multicolumn{2}{c}{\textbf{JSON}} & \multicolumn{2}{c}{\textbf{Images}} & Confidence\\
\cmidrule(lr){3-4} \cmidrule(lr){5-6} \cmidrule(lr){7-8}
\textbf{Strategy} & \textbf{Model} & \textbf{Internal} & \textbf{Verbal} & \textbf{Internal} & \textbf{Verbal} & \textbf{Internal} & \textbf{Verbal} & \textbf{Gap}\\
\midrule
\multirow{4}{*}{Zero Shot} 
& Qwen3-VL & 81.36 & 60.8 & 71.55 & 48.94 & 85.72 & 80.94 & 15.98\\
& Gemma3 & 83.74 & 86.30 & 75.23 & 82.11 & 83.50 & 85.98 & \textbf{3.98}\\
& Molmo2 & 53.53 & 64.8 & 38.9 & 43.2 & 45.84 & 96.22 & 21.98\\
& InternVL3.5 & 65.50 & 61.10 & 47.75 & 32.70 & 65.78 & 73.24 & 4.00\\
\midrule
\multirow{4}{*}{Few Shot} 
& Qwen3-VL & 77.4 & 52.6 & 60.2 & 25.58 & 83.58 & 77.02 & 21.99\\
& Gemma3 & 88.32 & 83.23 & 64.97 & 100.00 & 83.39 & 85.92 & \textbf{10.83}\\
& Molmo2 & 60.31 & 60.6 & 37.94 & 32.6 & 36.75 & 92.94 & 17.04\\
& InternVL3.5 & 67.79 & 48.54 & 43.49 & 26.47 & 69.54 & 48.86 & 18.98\\
\midrule
\multirow{4}{*}{CoT} 
& Qwen3-VL & 85.39 & 65.85 & 75.41 & 46.37 & 84.11 & 82.77 & 16.64\\
& Gemma3 & 88.66 & 82.83 & 67.95 & 82.82 & 82.81 & 85.54 & \textbf{3.92}\\
& Molmo2 & 56.67 & 55.38 & 28.25 & 41.56 & 49.53 & 90.99 & 17.82\\
& InternVL3.5 & 77.74 & 61.16 & 66.66 & 30.65 & 83.99 & 74.60 & 20.66\\
\midrule
\multicolumn{2}{l}{\textbf{Average}} & \textcolor{ForestGreen}{\textbf{73.87}} & \textcolor{red}{{65.26}} & \textcolor{ForestGreen}{\textbf{56.525}} & \textcolor{red}{{49.41}} & \textcolor{red}{71.21} & \textcolor{ForestGreen}{\textbf{81.26}} & --\\
\bottomrule
\end{tabular}
\end{table}

\subsection{Conformal Prediction for Uncertainty Quantification}
\label{sec:uncert_quant}
While attribution metrics evaluate whether a model's cited cell supports the answer, they do not quantify how uncertain the response is. Therefore, we study uncertainty quantification (UQ) as presented by \citep{ye2024benchmarkingllmsuncertaintyquantification}. We employ split-conformal prediction, which converts the models per-cell confidence scores to a prediction set $C(x)$ which uses a user-controlled target error rate $\alpha$. 

\paragraph{Setup and Adaptations:}
We utilize the standard split-conformal partition of calibration ($\mathcal{D}_{\text{cal}}$) and test ($\mathcal{D}_{\text{test}}$) sets. However, we introduce two specific adaptations to handle the nature of generative table attribution:

\begin{itemize}
    \item \textbf{Multi-Cell Coverage:} Unlike standard classification where the label is a single token, a ground truth answer in our task may span a region of cells. We therefore define the coverage criterion as satisfied if \textit{any} ground truth cell is present in the predicted set $C(x)$.
    \item \textbf{Open-Vocabulary Approximation:}  Since our model operates over an open vocabulary, and not a fixed label set, computing the normalizing constant over all possible table coordinates is computationally complex. We instead adopt a sparse approximation where probability mass is estimated only for the tokens actively generated by the model, assigning zero probability to non-generated coordinates.
\end{itemize}

\paragraph{Scoring Functions:}
We implement two non-conformity scoring functions, adapted from \citet{ye2024benchmarkingllmsuncertaintyquantification} to support our multi-cell coverage definition:

\begin{itemize}
    \item \textbf{Least Ambiguous Class (LAC):} This method constructs prediction sets based on absolute probability thresholds. We define the non-conformity score $s_i$ as one minus the probability of the \textit{most likely} correct cell:
    \begin{equation}
        s_i = 1 - \max_{y \in Y_{\text{true}}} \hat{p}(y \mid x)
    \end{equation}
    The prediction set is constructed by including all cells with probability $\hat{p}(y|x) \geq 1 - \hat{q}$, where $\hat{q}$ is the empirical quantile of scores over $\mathcal{D}_{\text{cal}}$.

    \item \textbf{Adaptive Prediction Sets (APS):} This method accumulates probability mass from the sorted predictions to account for the tail of the distribution. We define the non-conformity score $s_i$ as the minimum cumulative mass required to reach \textit{any} valid ground truth cell:
    \begin{equation}
        s_i = \min_{y \in Y_{\text{true}}} A(y)
    \end{equation}
    where $A(y)$ is the cumulative probability mass of candidate cells sorted in descending order. The prediction set includes candidates until the cumulative mass exceeds the calibrated threshold $\hat{q}$.
\end{itemize}

\begin{table}[t]
\centering
\footnotesize
\caption{Model uncertainty quantification results for attribution across representations and prompting strategies. We report average prediction set size (SS, number of cells) and coverage rate (CR, \%).}
\label{tab:attribution_uq_4b}
\begin{tabular}{@{}llcccccc@{}}
\toprule
& & \multicolumn{2}{c}{\textbf{Markdown}} & \multicolumn{2}{c}{\textbf{JSON}} & \multicolumn{2}{c}{\textbf{Images}} \\
\cmidrule(lr){3-4} \cmidrule(lr){5-6} \cmidrule(lr){7-8}
\textbf{Strategy} & \textbf{Model}
& \textbf{SS} & \textbf{CR}
& \textbf{SS} & \textbf{CR}
& \textbf{SS} & \textbf{CR} \\
\midrule
\multirow{4}{*}{Zero Shot}
& Qwen3-VL    & 184.96 & 83.50 & 186.58 & 79.50 & 187.22 & 87.50 \\
& Gemma3   & 165.61 & 80.32 & 183.73 & 80.72 & 186.49 & 83.70 \\
& Molmo2     & 184.11 & 83.50 & 187.80 & 79.00 & 187.53 & 90.10 \\
& InternVL3.5 & 183.09 & 80.50 & 185.54 & 81.50 & 177.33 & 79.00 \\
\midrule
\multirow{4}{*}{Few Shot}
& Qwen3-VL    & 183.67 & 82.50 & 188.79 & 81.00 & 188.16 & 87.30 \\
& Gemma3   & 171.83 & 80.12 & 182.26 & 80.12 & 186.48 & 86.80 \\
& Molmo2     & 187.31 & 85.00 & 185.07 & 78.00 & 188.16 & 83.10 \\
& InternVL3.5 & 174.46 & 81.50 & 177.44 & 79.00 & 177.33 & 79.00 \\
\midrule
\multirow{4}{*}{Chain of Thought}
& Qwen3-VL    & 173.60 & 84.00 & 179.17 & 80.50 & 184.89 & 88.80 \\
& Gemma3   & 183.27 & 79.32 & 180.59 & 78.51 & 185.97 & 83.70 \\
& Molmo2     & 181.05 & 81.50 & 188.55 & 80.50 & 184.55 & 83.50 \\
& InternVL3.5 & 185.15 & 77.00 & 187.21 & 78.50 & 186.40 & 77.00 \\
\bottomrule
\end{tabular}
\end{table}

\paragraph{Results:} Table 5 reports uncertainty quantification results across representations and prompting strategies using prediction set (SS) and coverage rate (CR). Across all representations and prompting methods, coverage rates remain close to the target level, while prediction set sizes vary substantially by modality, with image based inputs consistently producing larger sets than markdown and JSON. This indicates higher attribution uncertainty under visual perturbations.  

\subsection{Complete Attribution Metrics for All Model Families}
\label{app:full_results}
Table~\ref{app:full_results_gemma},~\ref{app:full_results_qwen},~\ref{app:full_results_molmo}, and~\ref{app:full_results_internvl} report the attribution metrics scores for the Gemma3, Qwen3-VL, Molmo2, and InternVL3.5 model families respectively.

\begin{table}[t]
\centering
\small
\caption{Attribution Metrics for Gemma Model Family}
\label{app:full_results_gemma}
\begin{tabular}{@{}llcccccc@{}}
\toprule
& & \multicolumn{2}{c}{\textbf{Markdown}} & \multicolumn{2}{c}{\textbf{JSON}} & \multicolumn{2}{c}{\textbf{Images}} \\
\cmidrule(lr){3-4} \cmidrule(lr){5-6} \cmidrule(lr){7-8}
\textbf{Strategy} & \textbf{Model} & \textbf{F1} & \textbf{Accuracy} & \textbf{F1} & \textbf{Accuracy} & \textbf{F1} & \textbf{Accuracy} \\
\midrule
\multirow{3}{*}{Zero Shot} 
& Gemma3-4b  & 0.159 & 13.40 & 0.113 & 0.80 & 0.887 & 16.60 \\
& Gemma3-12b & 0.306 & 27.20 & 0.141 & 0.60 & 0.396 & 37.10 \\
& Gemma3-27b & 0.372 & 31.00 & 0.279 & 1.51 & 0.464 & 44.10 \\
\midrule
\multirow{3}{*}{Few Shot} 
& Gemma3-4b  & 0.713 & 6.20 & 0.006 & 0.60 & 0.160 & 14.20 \\
& Gemma3-12b & 0.247 & 22.80 & 0.004 & 0.40 & 0.396 & 36.90 \\
& Gemma3-27b & 0.279 & 26.50 & 0.002 & 0.00 & 0.393 & 37.13 \\
\midrule
\multirow{3}{*}{Chain of Thought} 
& Gemma3-4b  & 0.11 & 10.00 & 0.0036 & 0.20 & 0.16 & 14.10 \\
& Gemma3-12b & 0.26 & 24.20 & 0.0087 & 0.60 & 0.42 & 39.90 \\
& Gemma3-27b & 0.47 & 44.50 & 0.0125 & 1.00 & 0.47 & 45.90 \\
\midrule
\multicolumn{2}{l}{\textbf{Average}} 
& 0.253 & 22.87 & 0.01 & 0.63 & 0.340 & 31.77 \\
\bottomrule
\end{tabular}
\end{table}

\begin{table}[t]
\centering
\small
\caption{Attribution Metrics for Qwen3-VL Model Family}
\label{app:full_results_qwen}
\begin{tabular}{@{}llcccccc@{}}
\toprule
& & \multicolumn{2}{c}{\textbf{Markdown}} & \multicolumn{2}{c}{\textbf{JSON}} & \multicolumn{2}{c}{\textbf{Images}} \\
\cmidrule(lr){3-4} \cmidrule(lr){5-6} \cmidrule(lr){7-8}
\textbf{Strategy} & \textbf{Model} & \textbf{F1} & \textbf{Accuracy} & \textbf{F1} & \textbf{Accuracy} & \textbf{F1} & \textbf{Accuracy} \\
\midrule
\multirow{4}{*}{Zero Shot} 
& Qwen3-VL-2b  & 0.10  & 8.50  & 0.005 & 0.50 & 0.15 & 14.10 \\
& Qwen3-VL-4b  & 0.379 & 35.50 & 0.015 & 1.00 & 0.48 & 45.40 \\
& Qwen3-VL-8b  & 0.261 & 23.50 & 0.004 & 0.00 & 0.40 & 37.50 \\
& Qwen3-VL-32b & 0.515 & 49.50 & 0.021 & 1.00 & 0.72 & 69.70 \\
\midrule
\multirow{4}{*}{Few Shot} 
& Qwen3-VL-2b  & 0.068 & 6.50  & 0.010 & 1.00 & 0.09 & 8.20 \\
& Qwen3-VL-4b  & 0.240 & 21.50 & 0.019 & 1.50 & 0.38 & 34.90 \\
& Qwen3-VL-8b  & 0.281 & 25.50 & 0.024 & 1.50 & 0.30 & 27.20 \\
& Qwen3-VL-32b & 0.390 & 36.00 & 0.003 & 0.00 & 0.69 & 68.10 \\
\midrule
\multirow{4}{*}{Chain of Thought} 
& Qwen3-VL-2b  & 0.220 & 20.00 & 0.003 & 0.00 & 0.32 & 29.80 \\
& Qwen3-VL-4b  & 0.515 & 49.00 & 0.016 & 1.00 & 0.47 & 44.40 \\
& Qwen3-VL-8b  & 0.570 & 52.50 & 0.001 & 0.00 & 0.68 & 64.80 \\
& Qwen3-VL-32b & 0.685 & 59.00 & 0.018 & 1.00 & 0.77 & 70.10 \\
\midrule
\multicolumn{2}{l}{\textbf{Average}} 
& 0.35 & 32.25 & 0.012 & 0.71 & 0.46 & 42.85 \\
\bottomrule
\end{tabular}
\end{table}

\begin{table}[t]
\centering
\small
\caption{Attribution Metrics for Molmo2 Model Family}
\label{app:full_results_molmo}
\begin{tabular}{@{}llcccccc@{}}
\toprule
& & \multicolumn{2}{c}{\textbf{Markdown}} & \multicolumn{2}{c}{\textbf{JSON}} & \multicolumn{2}{c}{\textbf{Images}} \\
\cmidrule(lr){3-4} \cmidrule(lr){5-6} \cmidrule(lr){7-8}
\textbf{Strategy} & \textbf{Model} & \textbf{F1} & \textbf{Accuracy} & \textbf{F1} & \textbf{Accuracy} & \textbf{F1} & \textbf{Accuracy} \\
\midrule
\multirow{2}{*}{Zero Shot} 
& Molmo2-4b  & 0.218  & 19.5  & 0.013 & 1.00 & 0.355 & 33.6 \\
& Molmo2-8b  & 0.222 & 20.00 & 0.015 & 0.05 & 0.369 & 34.80 \\
\midrule
\multirow{2}{*}{Few Shot} 
& Molmo2-4b  & 0.196 & 18.00  & 0.01 & 0.50 & 0.215 & 20.20 \\
& Molmo2-8b  & 0.096 & 8.50 & 0.00 & 0.00 & 0.243 & 22.80 \\
\midrule
\multirow{2}{*}{Chain of Thought} 
& Molmo2-4b  & 0.162 & 15.50 & 0.019 & 1.00 & 0.243 & 22.90 \\
& Molmo2-8b  & 0.205 & 19.50 & 0.013 & 0.50 & 0.3376 & 31.70 \\
\midrule
\multicolumn{2}{l}{\textbf{Average}} 
& 0.183 & 16.83 & 0.011 & 0.58 & 0.294 & 27.66 \\
\bottomrule
\end{tabular}
\end{table}

\begin{table}[t]
\centering
\small
\caption{Attribution Metrics for InternVL3.5 Model Family}
\label{app:full_results_internvl}
\begin{tabular}{@{}llcccccc@{}}
\toprule
& & \multicolumn{2}{c}{\textbf{Markdown}} & \multicolumn{2}{c}{\textbf{JSON}} & \multicolumn{2}{c}{\textbf{Images}} \\
\cmidrule(lr){3-4} \cmidrule(lr){5-6} \cmidrule(lr){7-8}
\textbf{Strategy} & \textbf{Model} & \textbf{F1} & \textbf{Accuracy} & \textbf{F1} & \textbf{Accuracy} & \textbf{F1} & \textbf{Accuracy} \\
\midrule
\multirow{4}{*}{Zero Shot} 
& InternVL3.5-4b  & 0.273  & 42.50  & 0.013 & 1.50 & 0.441 & 53.10 \\
& InternVL3.5-8b  & 0.193 & 17.50 & 0.016 & 1.00 & 0.254 & 22.50 \\
& InternVL3.5-14b  & 0.44 & 46.50 & 0.026 & 1.50 & 0.557 & 56.30 \\
& InternVL3.5-38b & 0.595 & 56.50 & 0.023 & 1.50 & 0.616 & 59.40 \\
\midrule
\multirow{4}{*}{Few Shot} 
& InternVL3.5-4b  & 0.125 & 36.50  & 0.008 & 1.00 & 0.136 & 52.60 \\
& InternVL3.5-8b  & 0.190 & 16.50 & 0.015 & 1.00 & 0.277 & 25.30 \\
& InternVL3.5-14b  & 0.375 & 37.00 & 0.013 & 0.50 & 0.533 & 57.30 \\
& InternVL3.5-38b & 0.448 & 42.50 & 0.013 & 1.00 & 0.541 & 52.50 \\
\midrule
\multirow{4}{*}{Chain of Thought} 
& InternVL3.5-4b  & 0.364 & 39.00 & 0.01 & 0.50 & 0.534 & 54.70 \\
& InternVL3.5-8b  & 0.379 & 36.00 & 0.017 & 1.00 & 0.49 & 45.80 \\
& InternVL3.5-14b  & 0.447 & 40.50 & 0.018 & 1.00 & 0.582 & 55.40 \\
& InternVL3.5-38b & 0.607 & 58.00 & 0.003 & 0.00 & 0.646 & 59.5 \\
\midrule
\multicolumn{2}{l}{\textbf{Average}} 
& 0.369 & 39.08 & 0.014 & 0.95 & 0.467 & 49.53\\
\bottomrule
\end{tabular}
\end{table}

\end{document}